# Accurate Tumor Tissue Region Detection with Accelerated Deep Convolutional Neural Networks

Gabriel Tjio*, Xulei Yang*, Jia Mei Hong, Sum Thai Wong, Vanessa Ding, Andre Choo and Yi Su

Agency for Science, Technology and Research (A*STAR), Singapore

Manual annotation of pathology slides for cancer diagnosis is laborious and repetitive. Therefore, much effort has been devoted to develop computer vision solutions. Our approach, (FLASH), is based on a Deep Convolutional Neural Network (DCNN) architecture. It reduces computational costs and is faster than typical deep learning approaches by two orders of magnitude, making high throughput processing a possibility. In computer vision approaches using deep learning methods, the input image is subdivided into patches which are separately passed through the neural network. Features extracted from these patches are used by the classifier to annotate the corresponding region. Our approach aggregates all the extracted features into a single matrix before passing them to the classifier. Previously, the features are extracted from overlapping patches. Aggregating the features eliminates the need for processing overlapping patches, which reduces the computations required. DCCN and FLASH demonstrate high sensitivity (~ 0.96), good precision (~0.78) and high F1 scores (~0.84). The average time taken to process each sample for FLASH and DCNN is 96.6 seconds and 9489.20 seconds, respectively. Our approach was approximately 100 times faster than the original DCNN approach while simultaneously preserving high accuracy and precision.

## 1 Background

Kidney cancer accounts for 2-3% cancer cases globally, with 338,000 newly diagnosed cases and 144,000 deaths reported in 2012 (1). Clear cell renal cell carcinoma (CCRCC), a major histological subtype of kidney cancer, makes up 80% of malignant tumors in the kidneys (2). Although 5-year survival rates are high (70-80%) for Stage I and II patients, survival rates are only 50% for Stage III patients, and 8% for Stage IV patients (3).

To improve cancer treatment, there has been an increased effort in biomarker discovery for both new therapeutic targets and better diagnosis (4). As part of the biomarker discovery efforts, immunohistochemistry (IHC) technique has been routinely utilized to screen for proteins differentially expressed in normal and cancerous tissues (5). In addition, tissue microarrays (TMAs) have been increasingly employed (5, 6) to enable simultaneous screening of large quantity of samples. TMA is an array of multiple tissue samples arranged in a grid on a single microscope slide. Each sample, or tissue core, has a diameter ranging from 0.6 to 2.5 mm, and contains sufficient histological information representative of the original tissue specimen. TMAs allow numerous samples to be concurrently immune-stained under identical conditions for unbiased comparison. Conventionally, pathologists analyze IHC by scoring the chromogenic stain intensities, i.e., 3,3'-diaminobenzidine (DAB), in the tumor cells. However, low throughput associated with manual visual evaluations (7) creates bottlenecks in the translational research process.

Computational methods for quantitative analysis of IHC have been developed to reduce subjectivity and to eliminate prohibitively time consuming work for pathologists (8, 9). Possible tasks in digital pathology that could also be automated include nuclei detection, cell classification, detection of multicellular structures and tissue segmentation (10). Despite technological advances in automated tissue annotation, many commercially available algorithms still require tumor regions to be manually annotated before further quantitative analysis of IHC can be done (11, 12). This is because conventional computational approaches face difficulties in accurately identifying tumor tissue, due to the heterogeneity of cells that make up the tumor tissue (13).

---

*These authors have contributed equally in this work.

In this paper, we demonstrate a novel accelerated deep learning approach to detect tumor tissue regions from histopathology core images stained with 3,3'-diaminobenzidine and counterstained with hematoxylin. Typically, deep learning approaches utilize the sliding window technique to extract image patches. The class label for the pixel centered at the image patch is determined given the context from the image patch. Our accelerated deep learning approach addresses the oversampling inherent in the sliding window technique by aggregating the detected features from all the image patches during detection. The novelty of our proposed approach lies in the modified sliding window approach applied during detection. This modified sliding window accelerates computations while maintaining good performance. The two key modifications are 1) employing a sliding window with increased stride length to eliminate overlap between extracted patches, and 2) introducing a second sliding window to extract features from feature space instead of image space.

Generally, the critical steps for detection problems in computer vision involve preprocessing, feature extraction and either segmentation or classification. Feature extraction is particularly important because it ultimately determines classification performance (14). For supervised machine learning methods, a training set of labelled data is required for feature extraction. Jafari-Khouzani and Soltanina Zadeh demonstrated 97% accuracy in grading of prostate cancer from pathological images by using multiwavelet transforms to extract distinctive features (15). Vanderbeck *et al.* applied support vector machines to identify white regions from liver biopsy images and demonstrated overall accuracy of 89% (16).

Recently, however, Deep Convolutional Neural Networks (DCNN) outperformed other computer vision methods by a large margin (17) and have since become the focus of much research. The mathematical framework of DCNN is briefly discussed(18). DCNN has been applied to solve detection problems in pathology (19, 20) and it is primarily based on a model of visual perception (21) that suggests "simple cells" detect basic details (such as edges and corners). The output from several "simple cells" is then combined by "complex cells" to form more complex outputs. Similarly, in DCNN, the initial layers detect basic features which are subsequently combined by the following layers to form more complex representations. This enables DCNN to detect high level, abstract features from images, making it well suited for computer vision applications. Another advantage of Deep Convolutional Neural Networks over other forms of machine learning such as support vector machines is that deep convolutional neural networks do not require feature engineering and instead identifies features from a labelled training set.

The neural network is trained by passing a series of image patches, each with a class label to the network. Given an unlabelled image patch, the neural network can predict the label for the central pixel of that patch. Typically, an image is subdivided into multiple overlapping patches using the sliding window technique and passed to the neural network.

One limitation of DCNN, however, is the high computational resources required for training and detection, which limit the speed of training and implementing neural networks. Past work on accelerating deep learning methods can be classified into the following categories: use of transforms, such as Fast Fourier transforms, that simplify convolution and speed up computation (22), and methods which reduce redundancy in the neural networks (23).

Our accelerated tumour detection approach is closely related to the approach used by Hou *et al.* for classification of whole slide tissue images (24). Hou *et al.* proposed the use of a patch level classifier instead of an image level classifier because of the difficulties involved in training a deep convolutional neural network for high resolution whole slide images. The patch level classifier was applied to classify whole slide images. The separate output from these classifiers was then passed to a multi-class logistic regression to provide an image level classification. Similarly, our approach is trained on multiple patches and during evaluation, our output from the different patches are combined to provide an overall classification for the tissue core image. Rather than combining the output using a multi class logistic regression, our approach aggregates all the extracted features before classification into a single matrix. This eliminates redundant computations because there is no need to duplicate computations for overlapping patches extracted using the sliding window approach.

## 2 Methods

### 2.1 Sample Preparation

A 96-core kidney cancer TMA was procured from Pantomics, Inc. (KIC962; clear cell carcinoma cores n = 30 from 15 patients; mixed clear cell and papillary renal cell carcinoma n = 2 from 1 patient). The slide was baked in a 60°C oven for 30 minutes before deparaffinization and rehydration by washing the slide with xylene and decreasing percentages of ethanol in water. Antigens were retrieved by incubating the slide in 10 mM sodium citrate buffer with 0.05% Tween-20, pH 6, for 15 minutes at 95°C. The slide was subsequently blocked in 10% normal goat serum in PBS (Phosphate Buffered Saline), and incubated with an in-house kidney cancer-specific antibody overnight at 4°C. Next, endogenous peroxidase was quenched with 1% hydrogen peroxide in PBS for 30 minutes, followed by labeling the tissue with horseradish peroxidase (HRP)-conjugated secondary antibody polymer (Dako). The chromogenic stain was then developed with 3,3'-diaminobenzidine (DAB; Dako), and section counterstained with hematoxylin. Lastly, the TMA was dehydrated through ascending percentages of ethanol in water and xylene before mounting in acrylic resin (Sigma-Aldrich). Whole slide image was acquired with Leica SCN400 slide scanner at 0.5 μm per pixel (20× objective lens). Out of the 32 clear cell cores, 11 cores were found to not contain tumor clear cells, and were therefore excluded from our study.

### 2.2 Data Augmentation

The training data in this study are identical to the data used in an earlier study (25). The data were augmented to increase the amount of data available for training. 577 regions of dimensions 300×300 were extracted from kidney cancer tissue microarray cores. 243 regions are non-tumor tissue regions and 334 regions are tumor regions. All regions were rotated (-180°, -150°, -120°, -90°, -60°, -30°, 30°, 60°, 90°, 120°, 150°, 180°), cropped and resized to generate 21,349 image patches of dimensions 32×32.

### 2.3 Convolutional Neural Network

The neural network architecture for DCNN and FLASH are summarized in Table . For the convolutional layers, Glorot initialization was implemented with the weights sampled from a uniform distribution. The activation function used was the Rectified Linear Units (ReLUs). Both DCNN and FLASH were trained identically and the training workflow is described in Figure 1.

The network were trained using stochastic gradient descent with Nesterov momentum=0.9 and learning rate=0.002 for 500 epochs. Mirror padding was used to ensure that the inputs to a given layer have the same dimensions as the outputs. A dropout of 0.5 was implemented in the fully connected layer and the output layer to minimize overfitting. The weights in the DCNN model were updated using the stochastic gradient descent backpropagation approach. Distinct features which can consistently distinguish between the classes (i.e., tumor and non-tumor) were automatically learnt during the entire training step.

Table 1: Neural network architecture for the DCNN and the proposed approach FLASH.

| Layer | Depth | Kernel | Stride | Spatial Size | Parameters |
|-------|-------|--------|--------|--------------|------------|
| Input | 3 | ---- | ---- | 32×32×3 | 0 |

| Convolution | 16 | 3×3 | 1×1 | 32×32×16 | 3×3×3×16 |
| --- | --- | --- | --- | --- | --- |
| Pooling | 16 | 2×2 | 2×2 | 16×16×16 | 0 |
| Convolution | 32 | 3×3 | 1×1 | 16×16×32 | 3×3×16×32 |
| Pooling | 32 | 2×2 | 2×2 | 8×8×32 | 0 |
| Convolution | 64 | 3×3 | 1×1 | 8×8×64 | 3×3×32×64 |
| Pooling | 64 | 2×2 | 2×2 | 4×4×64 | 0 |
| Fully Connected | 256 | ---- | ---- | 1×1×256 | 4×4×64×256 |
| Output | 2 | ---- | ---- | 1×1×2 | 256×2 |

FLASH and DCNN were implemented using Theano (26) and Lasagne with the Python programming language. Both models were run on a Intel(R) Core(TM) i7-5820K workstation, 6 CPU cores, 3.30GHz, with a GeForce GTX 980 GPU.

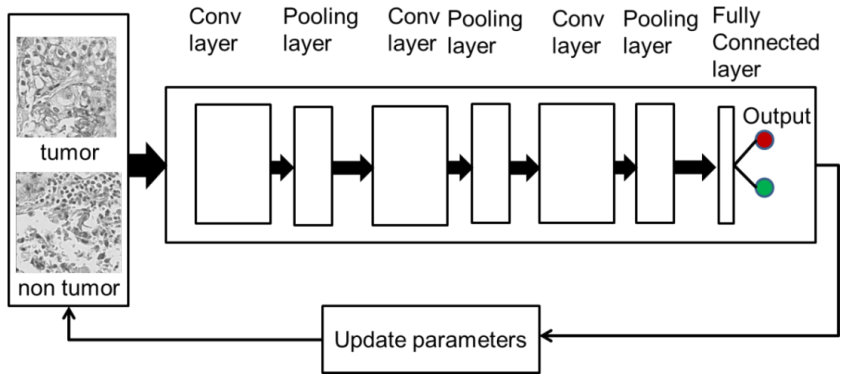

Figure 1: Training the DCNN model. The labeled training data (i.e., tumor and non-tumor) are passed to the convolutional (conv) layers, pooling layers and fully connected layers in the DCNN model. The output from the DCNN model is evaluated against the ground truth. The parameters in the DCNN model are updated iteratively.

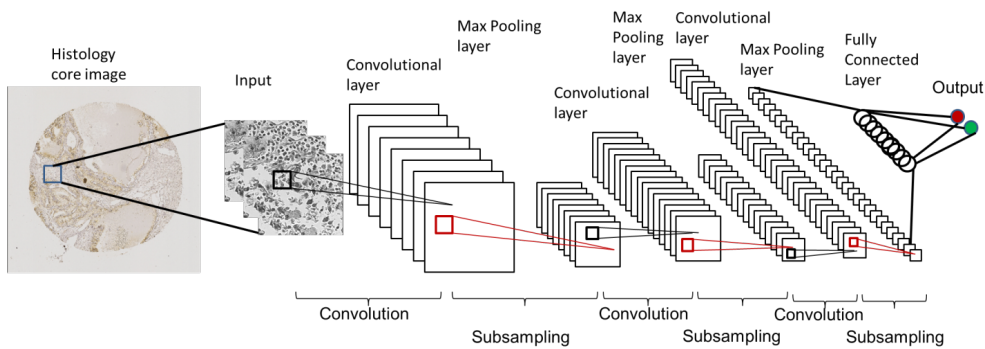

Figure 2: Original workflow for DCNN.

As the sliding window transverses the histological core, the region bounded by the window is passed to the DCNN model.
The output depicts the distribution of tumor/non tumor regions within the receptive field. This step is repeated until the entire histological core has been evaluated.

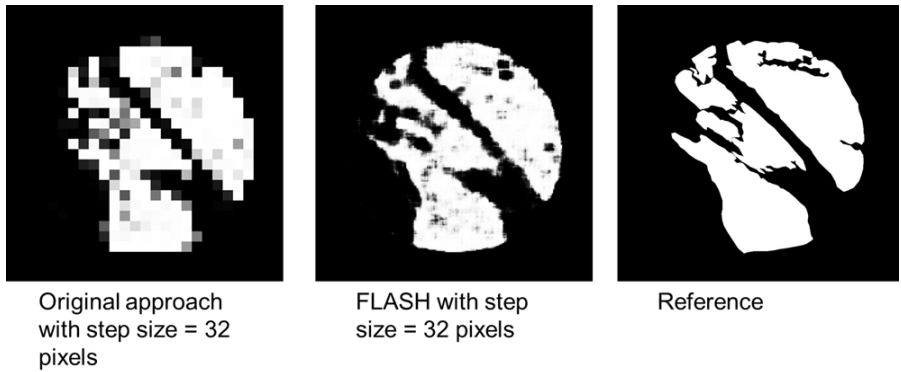

Figure 3: Illustration of the effect of sliding window stride lengths on the original DCNN approach and our proposed FLASH method. The reference image is obtained from manual annotations performed by a trained pathologist.

The key contribution of our proposed approach is that our modified sliding window approach allows for an increased stride length without reducing performance. Every image patch passed to the neural network has an output label corresponding to the central pixel of that patch. Therefore, the sliding window shifts by a pixel so that every pixel in the image is classified. Increasing the stride length would mean that some pixels in the original image would be passed over for processing. The pixels in the output image corresponding to the 'skipped' pixels would have to be derived via interpolation from neighboring pixels. Merely increasing the step size of the sliding window would result in a considerable loss of information due to lower resolution of the output image. Figure 3 illustrates that increasing the stride length results in a severe drop in performance in the original DCNN approach.

Previously, the DCNN approach evaluated the entire image one pixel at a time by shifting the receptive field with a step size of one pixel per iteration (25). However, this oversamples the input image and increases computation time. Our approach (FLASH), as illustrated in Figure 4, addresses this problem by modifying the sliding window technique: The sliding window is shifted over the input image with a step size equal to the length of the sliding window. This minimizes the total computation time during implementation of the DCNN approach and prevents oversampling of the input image.

Feature extraction was performed in feature space instead of image space like in DCNN. To illustrate our approach, let us consider a RGB square image with dimensions: $l \times l \times 3$ where $l$ is the width of the image and 3 refers to the number of color channels. The dimensions of the sliding window are $w \times w \times 3$. The number of patches extracted for a typical sliding window approach for DCNN is:

Number of patches with typical approach = $(l - w + 1)^2$

In contrast to typical sliding window approaches, our approach uses a larger step size, which avoids any overlap between the extracted patches.

Number of patches with our approach = $\left(floor\left(\frac{l}{w}\right)\right)^2$

For a sliding window of $w > 1$ pixel and $l > w$, the total number of patches extracted for our approach will be much less than the number of patches extracted for a typical sliding window approach, reducing the total number of computations required during detection. The features from the last convolutional layer from each patch were aggregated in a single matrix.

A second sliding window of size 8×8×64 with a stride length of 1 pixel was used to extract the features from feature space. The features are subsequently passed to the pooling, fully connected and output layers. The output label was then assigned to the image pixels corresponding to the patch. This step was repeated until the entire image has been labeled.

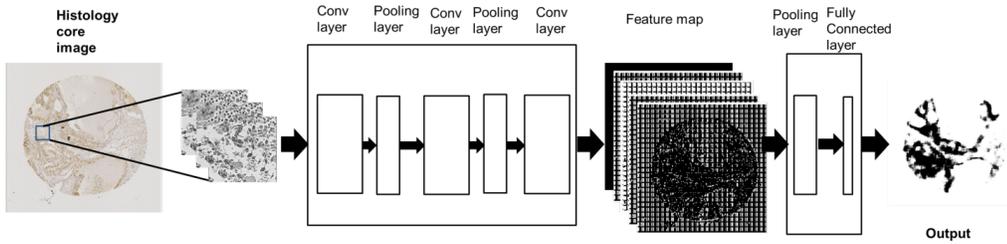

Figure 4: Workflow for FLASH.

*2.4 Evaluation*

21 TMA cores were manually annotated by a trained independent observer and used as the gold standard. The sensitivity, precision and F1 score of the original method and our modified approach were determined. Sensitivity refers to the proportion of tumour regions that are detected. Precision refers to the proportion of classified regions that were correctly identified. The F1 score provides a measure of the test accuracy and can be likened to a weighted average of the sensitivity and precision. Finally, specificity refers to the non-tumour regions that have been correctly classified.

$$\text{Sensitivity} = \frac{\text{True Positives}}{\text{True Positives} + \text{False Negatives}}$$

$$\text{Precision} = \frac{\text{True Positives}}{\text{True Positives} + \text{False Positives}}$$

$$F1 = \frac{2 \times \text{Sensitivity} \times \text{Precision}}{\text{Sensitivity} + \text{Precision}}$$

The Receiver Operator Characteristic (ROC) graphs were generated for both FLASH and DCNN. The area under the graph was determined using the trapezoidal formula. The calculated area is an estimate of the performance for both approaches.

3 Results

Table 2: Mean and standard deviations in sensitivity, precision and F1 scores (FLASH and DCNN).

|  | *Sensitivity* | *Precision* | *F1* |
| --- | --- | --- | --- |
| FLASH | $0.956 \pm 0.034$ | $0.775 \pm 0.214$ | $0.837 \pm 0.187$ |
| DCNN | $0.955 \pm 0.032$ | $0.774 \pm 0.214$ | $0.836 \pm 0.186$ |

The mean and standard deviation of sensitivity, precision and F1 scores (FLASH and DCNN) are shown in Table .

Table 3: Comparison of mean and standard deviation in processing time for FLASH and DCNN.

|  | FLASH | DCNN |
| --- | --- | --- |

| Processing time (s) | 96.65 ± 3.96 | 9489.2 ± 115.57 |

Table shows the mean and standard deviation in processing time for FLASH and DCNN.

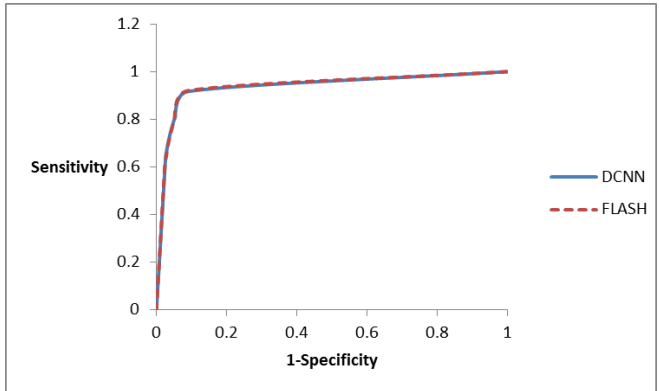

Figure 5: Receiver operator characteristic (ROC) graphs for original and FLASH.

From Figure 5, the area under the ROC graph for the original DCNN approach is 0.932 while the area under the graph for the FLASH approach is 0.934. Both graphs also closely overlap each other.

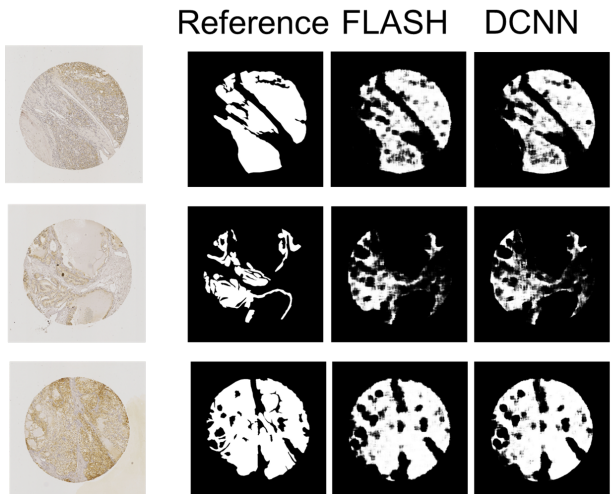

Figure 6: Output from DCNN and FLASH for selected histopathology core images compared with the corresponding manually annotated images (reference).

Figure 6 shows the output from DCNN and FLASH together with the manually annotated output images.

## 4 Conclusion

Both approaches demonstrate high sensitivity (~ 0.96), good precision (~0.78) and F1 scores (~0.84). The results suggest that our approach preserves the accuracy and precision while accelerating the tumour detection process by almost 100 times. The average sensitivity, precision and F1 scores for both FLASH and DCNN were similar. The area under the graph for the ROC is a measure of accuracy for the test and both approaches show equivalent performance. Additionally, qualitative comparison of the output (Figure 6) from DCNN and FLASH show that there is no performance drop in FLASH despite the modifications.

One limitation of our approach is the downsampling of the image during pooling. This can result in some loss of detail for fine structures. This could be a problem in tissue samples comprising of large quantities of dispersed, scattered cells. There is some trade-off between speedup on detection time and the output resolution, though we show that the downsampling does not significantly affect the quantitative and qualitative results.

Our method can be further accelerated by restricting the detection steps to regions that contain cells. For instance, Ren et al. demonstrated the use of region proposal networks to accelerate object detection (27). This approach will also likely improve computational speeds for tissue cores that contain few cells by limiting computations to certain regions.

Future work will focus on improving the accuracy of deep learning methods in classifying these fine structures. Fully convolutional neural networks have been explored as a means of improving spatial precision in the computed output (28). In another example, Ronneberger *et al.* recently demonstrated a deep learning method that achieves high localization accuracy (29) and this would be another possible means of extending the work described in this study.

## Acknowledgment

This work is supported by the research grant (Grant No. 1431AFG125) under the Joint Council Office (JCO), Agency for Science, Technology and Research (A*STAR), Singapore.

## References

1. Ferlay J, Soerjomataram I, Dikshit R, Eser S, Mathers C, Rebelo M, et al. Cancer incidence and mortality worldwide: sources, methods and major patterns in GLOBOCAN 2012. International journal of cancer. 2015;136(5):E359-86.
2. Cheville JC, Lohse CM, Zincke H, Weaver AL, Blute ML. Comparisons of outcome and prognostic features among histologic subtypes of renal cell carcinoma. The American journal of surgical pathology. 2003;27(5):612-24.
3. American_Cancer_Society. Survival rates for kidney cancer by stage [cited 2017 January 17]. Available from: http://www.cancer.org/cancer/kidneycancer/detailedguide/kidney-cancer-adult-survival-rates
4. Mabert K, Cojoc M, Peitzsch C, Kurth I, Souchelnytskyi S, Dubrovska A. Cancer biomarker discovery: current status and future perspectives. International journal of radiation biology. 2014;90(8):659-77.
5. Chiang SC, Han CL, Yu KH, Chen YJ, Wu KP. Prioritization of cancer marker candidates based on the immunohistochemistry staining images deposited in the human protein atlas. PloS one. 2013;8(11):e81079.
6. Hewitt SM. Tissue microarrays as a tool in the discovery and validation of predictive biomarkers. Methods in molecular biology (Clifton, NJ). 2012;823:201-14.
7. Borlot VF, Biasoli I, Schaffel R, Azambuja D, Milito C, Luiz RR, et al. Evaluation of intra- and interobserver agreement and its clinical significance for scoring bcl-2 immunohistochemical expression in diffuse large B-cell lymphoma. Pathology international. 2008;58(9):596-600.
8. Al-Kofahi Y, Lassoued W, Lee W, Roysam B. Improved Automatic Detection and Segmentation of Cell Nuclei in Histopathology Images. IEEE Transactions on Biomedical Engineering. 2010;57(4):841-52.
9. Simsek AC, Tosun AB, Aykanat C, Sokmensuer C, Gunduz-Demir C. Multilevel Segmentation of Histopathological Images Using Cooccurrence of Tissue Objects. IEEE Transactions on Biomedical Engineering. 2012;59(6):1681-90.

10. McCann MT, Ozolek JA, Castro CA, Parvin B, Kovacevic J. Automated Histology Analysis: Opportunities for signal processing. IEEE Signal Processing Magazine. 2015;32(1):78-87.
11. Rizzardi AE, Johnson AT, Vogel RI, Pambuccian SE, Henriksen J, Skubitz AP, et al. Quantitative comparison of immunohistochemical staining measured by digital image analysis versus pathologist visual scoring. Diagnostic pathology. 2012;7:42.
12. Keay T, Conway CM, O'Flaherty N, Hewitt SM, Shea K, Gavrielides MA. Reproducibility in the automated quantitative assessment of HER2/neu for breast cancer. Journal of pathology informatics. 2013;4:19.
13. Turley SJ, Cremasco V, Astarita JL. Immunological hallmarks of stromal cells in the tumour microenvironment. Nature reviews Immunology. 2015;15(11):669-82.
14. Gurcan MN, Boucheron LE, Can A, Madabhushi A, Rajpoot NM, Yener B. Histopathological Image Analysis: A Review. IEEE Reviews in Biomedical Engineering. 2009;2:147-71.
15. Jafari-Khouzani K, Soltanian-Zadeh H. Multiwavelet grading of pathological images of prostate. IEEE Transactions on Biomedical Engineering. 2003;50(6):697-704.
16. Vanderbeck S, Bockhorst J, Komorowski R, Kleiner DE, Gawrieh S. Automatic classification of white regions in liver biopsies by supervised machine learning. Human Pathology. 2014;45(4):785-92.
17. Krizhevsky A, Sutskever I, Hinton G. ImageNet Classification with Deep Convolutional Neural Networks. In: Pereira F, Burges CJC, Bottou L, Weinberger KQ, editors. Advances in Neural Information Processing Systems 25: Curran Associates, Inc.; 2012. p. 1097-105.
18. Mallat S. Understanding deep convolutional networks. Philosophical Transactions of the Royal Society A: Mathematical, Physical and Engineering Sciences. 2016;374(2065).
19. Cireşan DC, Giusti A, Gambardella LM, Schmidhuber J. Mitosis Detection in Breast Cancer Histology Images with Deep Neural Networks. In: Mori K, Sakuma I, Sato Y, Barillot C, Navab N, editors. Medical Image Computing and Computer-Assisted Intervention – MICCAI 2013: 16th International Conference, Nagoya, Japan, September 22-26, 2013, Proceedings, Part II. Berlin, Heidelberg: Springer Berlin Heidelberg; 2013. p. 411-8.
20. Sirinukunwattana K, Raza SEA, Tsang YW, Snead DRJ, Cree IA, Rajpoot NM. Locality Sensitive Deep Learning for Detection and Classification of Nuclei in Routine Colon Cancer Histology Images. IEEE Transactions on Medical Imaging. 2016;35(5):1196-206.
21. Hubel DH, Wiesel TN. Receptive fields, binocular interaction and functional architecture in the cat's visual cortex. The Journal of Physiology. 1962;160(1):106-54.2.
22. Lavin A, Gray S, editors. Fast Algorithms for Convolutional Neural Networks. 2016 IEEE Conference on Computer Vision and Pattern Recognition (CVPR); 2016 27-30 June 2016.
23. Denton E, Zaremba W, Bruna J, LeCun Y, Fergus R. Exploiting Linear Structure Within Convolutional Networks for Efficient Evaluation. ArXiv e-prints2014.
24. Hou L, Samaras D, Kurc TM, Gao Y, Davis JE, Saltz JH. Patch-based Convolutional Neural Network for Whole Slide Tissue Image Classification. Proceedings IEEE Computer Society Conference on Computer Vision and Pattern Recognition. 2016;2016:2424-33.
25. Yang X, Yeo SY, Hong JM, Wong ST, Tang WT, Wu ZZ, et al., editors. A deep learning approach for tumor tissue image classification. BioMed; 2016; Innsbruck, Austria
26. Al-Rfou R, Alain G, Almahairi A, Angermueller C, Bahdanau D, Ballas N, et al. Theano: A Python framework for fast computation of mathematical expressions. arXiv preprint arXiv:160502688. 2016.
27. Ren S, He K, Girshick R, Sun J. Faster R-CNN: Towards Real-Time Object Detection with Region Proposal Networks. IEEE transactions on pattern analysis and machine intelligence. 2017;39(6):1137-49.
28. Shelhamer E, Long J, Darrell T. Fully Convolutional Networks for Semantic Segmentation. IEEE Transactions on Pattern Analysis and Machine Intelligence. 2017;39(4):640-51.
29. Ronneberger O, Fischer P, Brox T. U-Net: Convolutional Networks for Biomedical Image Segmentation. In: Navab N, Hornegger J, Wells WM, Frangi AF, editors. Medical Image Computing and Computer-Assisted Intervention – MICCAI 2015: 18th International Conference, Munich, Germany, October 5-9, 2015, Proceedings, Part III. Cham: Springer International Publishing; 2015. p. 234-41.